\documentclass{article}

\usepackage{microtype}
\usepackage{graphicx}
\usepackage{booktabs} % for professional tables
\usepackage{enumitem}
\usepackage{fontawesome5}

\usepackage{subcaption}
\usepackage{hyperref}
\usepackage{multirow}
\usepackage{caption}
\usepackage{amsmath}
\usepackage{amssymb}
\usepackage{mathtools}
\usepackage{amsthm}
\usepackage{algorithm}
\usepackage{algpseudocode}
\usepackage{url}

\usepackage[accepted]{icml2025}

\usepackage[capitalize,noabbrev]{cleveref}

\theoremstyle{plain}

\theoremstyle{definition}

\theoremstyle{remark}

\usepackage[textsize=tiny]{todonotes}

\icmltitlerunning{Submission and Formatting Instructions for ICML 2025}

\begin{document}

\twocolumn[

\icmltitle{LiteLong: Resource-Efficient Long-Context Data Synthesis for LLMs}

\icmlsetsymbol{equal}{*}

\begin{icmlauthorlist}
\icmlauthor{Junlong Jia}{buaa}
\icmlauthor{Xing Wu}{iie,xhs}
\icmlauthor{Chaochen Gao}{iie}
\icmlauthor{Ziyang Chen}{iie}\\
\icmlauthor{Zijia Lin}{qinghua}
\icmlauthor{Zhongzhi Li}{xhs}
\icmlauthor{Weinong Wang}{xhs}
\icmlauthor{Haotian Xu}{xhs}
\icmlauthor{Donghui Jin}{buaa}
\icmlauthor{Debing Zhang}{xhs}
\icmlauthor{Binghui Guo}{buaa}

\end{icmlauthorlist}
\icmlaffiliation{buaa}{School of Artificial Intelligence, Beihang University}
\icmlaffiliation{iie}{Institute of Information Engineering, Chinese Academy of Sciences}
\icmlaffiliation{xhs}{Xiaohongshu Inc}
\icmlaffiliation{qinghua}{Tsinghua University}

% \icmlcorrespondingauthor{Chaochen Gao}{gaochaochen@iie.ac.cn}
\icmlcorrespondingauthor{Xing Wu}{wuxing@iie.ac.cn}

\icmlkeywords{Machine Learning, ICML}

\vskip 0.3in
]



\printAffiliationsAndNotice{}  % leave blank if no need to mention equal contribution

\begin{abstract}
    High-quality long-context data is essential for training large language models (LLMs) capable of processing extensive documents, yet existing synthesis approaches using relevance-based aggregation face challenges of computational efficiency. We present LiteLong, a resource-efficient method for synthesizing long-context data through structured topic organization and multi-agent debate. Our approach leverages the BISAC book classification system to provide a comprehensive hierarchical topic organization, and then employs a debate mechanism with multiple LLMs to generate diverse, high-quality topics within this structure. For each topic, we use lightweight BM25 retrieval to obtain relevant documents and concatenate them into 128K-token training samples. Experiments on HELMET and Ruler benchmarks demonstrate that LiteLong achieves competitive long-context performance and can seamlessly integrate with other long-dependency enhancement methods. LiteLong makes high-quality long-context data synthesis more accessible by reducing both computational and data engineering costs, facilitating further research in long-context language training.
\end{abstract}


\section{Introduction}

    Large language models (LLMs) have gained widespread attention due to their robust capabilities. Recently, LLM context windows have expanded significantly \cite{peng2024yarn,young2024yi,team2024qwen2,dots1}. The Llama series illustrates this trend, evolving from 4K tokens in Llama 2 \cite{touvron2023llama} to 128K in Llama 3.1 \cite{dubey2024llama}. This expanded capacity enables LLMs to address complex tasks like document summarization \cite{wu2023less}, question answering on books \cite{de2018question}, and Code Planning \cite{bairi2024codeplan}. Current approaches for modeling long-range dependencies typically continue training LLMs with documents reaching the target length. However, high-quality long documents remain scarce across most domains, creating a significant challenge as target context lengths increase \cite{gao2025quest}.

    Existing methods for long-context data synthesis are mainly relevance-based \cite{guu2020retrieval,levine2022the,shi2024incontext,gao2025quest}, which aggregate semantically relevant documents to form long-range dependencies. For example, ICLM \cite{shi2024incontext} constructs a semantic document graph via retrieval and indexing, and proposes a traveling salesman problem-based sorting algorithm to maximize contextual similarity while ensuring that each document is included only once. Similarly, Quest \cite{gao2025quest} uses a generative model to predict potential queries for each document, then groups documents with similar query patterns. While these methods improve document relevance, they face two significant limitations: (1) they require either generating embeddings or constructing queries over massive corpora, both of which demand substantial computational resources—the GPU cost incurred during the data synthesis process; and (2) they lack a clear framework to ensure diversity coverage in the generated content. This raises the question of \textit{whether we can achieve effective long-context data synthesis while maintaining resource efficiency}.

    \begin{figure*}[tbp]
        \centering
        \includegraphics[width=0.85\textwidth]{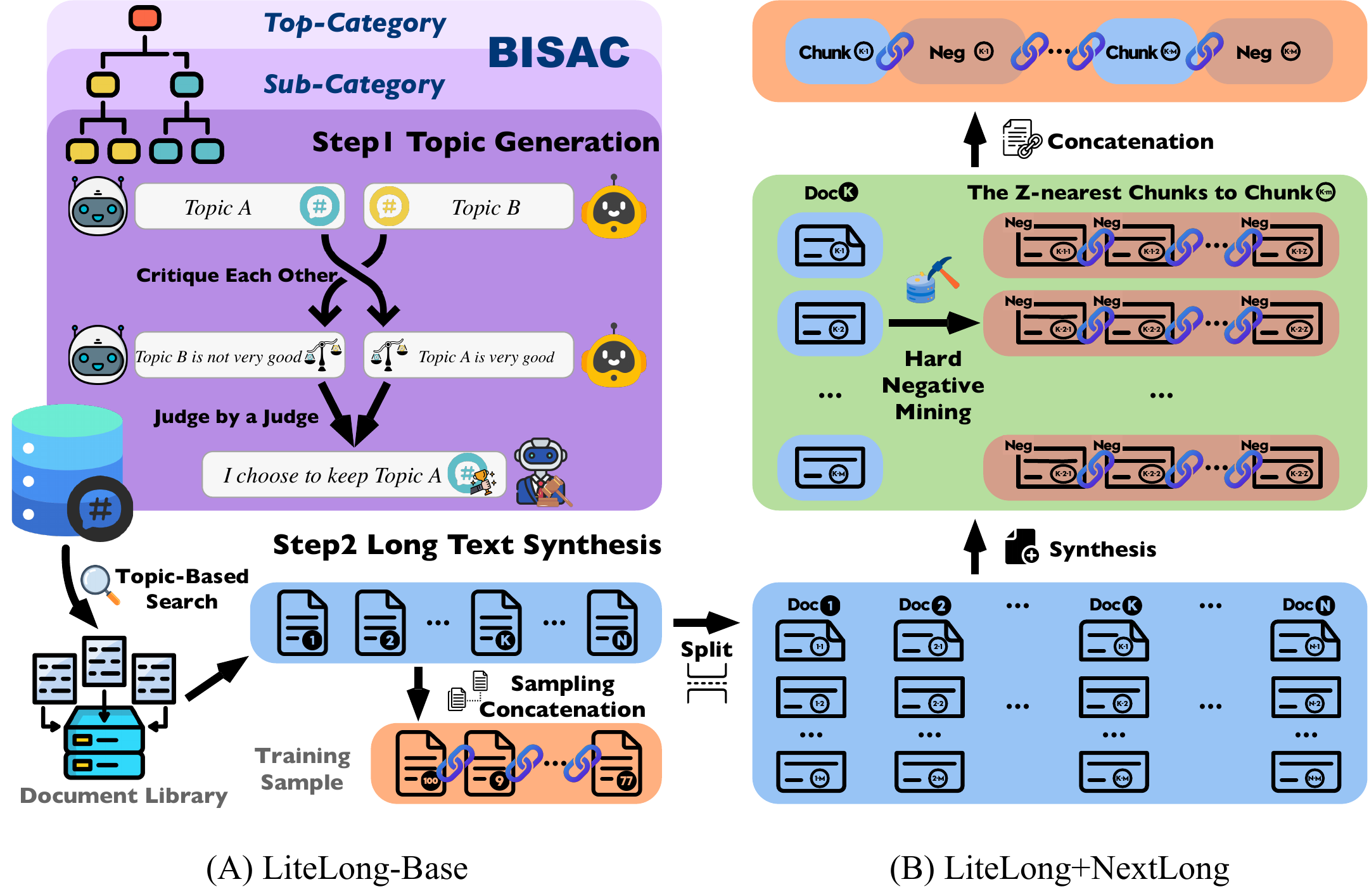}
        \caption{Overview of the LiteLong framework. The left subfigure illustrates the LiteLong method, focusing on topic synthesis using BISAC categories. The right subfigure shows the integration of LiteLong with the long-dependency enhancement method, NExtLong\cite{gao2025nextlong}.}
        \label{fig:method}
    \end{figure*}

    We draw inspiration from Cosmopedia V2 \cite{benallal2024smollmcorpus}, which adopts the Book Industry Standards and Communications (BISAC) \cite{martinez2016bisac} book classification—a comprehensive subject categorization system that offers better coverage and diversity than categories constructed through unsupervised clustering. The success of adopting BISAC provides a promising direction for creating large-scale long-context data without incurring high clustering computational costs as previous works \cite{shi2024incontext, gao2025quest}. 
    
    We start from the BISAC classification standard and use LLMs to generate diverse topics for each second-level category. We then employ lightweight BM25 \cite{robertson2009probabilistic} retrieval to obtain topic-relevant documents and concatenate them into long-context training data of the target length. This process is highly efficient, thus termed \textbf{LiteLong}, because the BISAC book classification system contains only a few thousand categories, minimizing GPU computation required for topic generation. 

    We further implement a multi-agent debate \cite{du2023improving, kenton2024on} mechanism to promote the diversity and quality of generated topics. This mechanism employs two separate Debate LLMs to independently generate topic candidates, then has them mutually critique each other's topics based on multiple criteria, such as topic relevance and diversity. Finally, the Judge LLM aggregates the critiques and adjudicates the best subset of topics for each secondary category. This mechanism has two main advantages: (1) the competitive generation process involves two independent LLMs, which can produce more diverse topics and avoid the limitations of a single LLM due to its training data; (2) the Judge LLM can reject weakly relevant topics, thereby improving the topic quality.

    The incorporation of the multi-agent debate mechanism does not significantly increase LiteLong's resource demands. Specifically, the debate is conducted at the topic level, rather than across full documents. Since the number of BISAC subcategories is only a few thousand, and each category only generates a limited number of topic candidates, the total number of LLM inference is relatively small. This contrasts sharply with relevance-based approaches that require expensive document embeddings or dense clustering over billions of tokens. 
    
    Through extensive experiments and evaluations on the HELMET \cite{yen2025helmet} and RULER \cite{hsieh2024ruler} benchmarks, we demonstrate that LiteLong achieves competitive performance in a resource-efficient setting. Furthermore, we find that LiteLong can seamlessly integrate with recently proposed long-dependency enhancement method NExtLong \cite{gao2025nextlong}, further improving the model's long-context capabilities. We also provide ablation studies investigating the key components of LiteLong. Overall, LiteLong provides a resource-efficient approach to long-context data synthesis, significantly reducing resource consumption during the data construction process, facilitating further research in long-context language modeling and contributing to the democratization of these capabilities.

    Our contributions are as follows:
    \begin{itemize}
        \item We present LiteLong, a resource-efficient method for synthesizing long-context data through structured topic organization and multi-agent debate mechanism.
        \item We demonstrate that LiteLong achieves competitive long-context performance and can be seamlessly integrated with recently proposed long-dependency enhancement method.
        \item We provide an in-depth analysis of the design choices in LiteLong to validate the effectiveness of each component.
    \end{itemize}

\section{Related Work}

\subsection{Long-Context Data Synthesis}

    The development of effective long-context language models is hindered by the scarcity of high-quality long-document training data. Several approaches are proposed to address this challenge. The simplest approach, known as Random Concatenation (Standard) \cite{ouyang2022training}, involves randomly joining shorter documents to create longer training samples \cite{xiong2023effective}. While computationally efficient, this method often produces incoherent transitions between documents, limiting the model's ability to develop robust long-range understanding. Similarity-based methods, such as KNN \cite{jiang2023longllmlingua}, aggregate documents based on content similarity, improving semantic coherence but potentially limiting exposure to diverse content patterns. Query-centric approaches, like Quest \cite{gao2025quest}, use a generative model to predict potential queries for each document, then group documents with similar query patterns. While effective, this approach introduces significant computational overhead for query generation. Document transformation methods, such as Untie the Knots (UtK) \cite{tian2024untie} and NExtLong \cite{gao2025nextlong}, manipulate document structure by chunking, shuffling, and incorporating negative examples to encourage the model to develop better document navigation capabilities, focusing primarily on improving attention mechanisms rather than semantic organization. LiteLong differentiates itself by leveraging an external topic framework to guide document organization, avoiding both the incoherence of random concatenation and the computational expense of query generation.

\subsection{Multi-Agent Collaboration in Content Generation}
    Recent research explores the potential of multi-agent systems for content generation and evaluation, highlighting several key approaches. Debate-based methods, such as those used in systems like Generative Agents \cite{park2023generative} and Debating AI \cite{irving2018ai}, utilize structured debates between multiple AI agents to enhance reasoning and content quality, showing particular promise in domains requiring complex evaluation criteria. Role-based collaboration methods, which employ specialized agent roles such as debater, critic, and editor, demonstrate improvements in content quality and diversity \cite{du2023improving, wu2023autogen}. Additionally, LLM agents are applied to synthetic data generation tasks, including instruction tuning data \cite{xu2024wizardlm} and reasoning examples \cite{wang2022self}. Our approach extends these ideas to the domain of long-context data synthesis, employing a novel debate mechanism between competing LLMs to ensure both diversity and quality in generated topics.

\section{Method}
    In this section, we describe LiteLong, our approach to efficient long-context data synthesis through multi-agent debate. We first introduce the BISAC \cite{martinez2016bisac} category system, and then detail the two main components of our method: (1) multi-agent debate topic generation, and (2) document retrieval and aggregation. Finally, we describe how LiteLong can be seamlessly integrated with NExtLong to further enhance long-range dependency modeling.

    \begin{algorithm}[!tbp]
        \caption{Efficient Long-Context Data Synthesis via Multi-Agent Debate}
        \label{alg:multi-agent}
        \begin{algorithmic}[1]
        % --- Notation description ---
        \State \textbf{Notation:}
        \State \hspace{1em} $\mathcal{T}_{\text{total}}$: All candidate topics generated by both Debate models
        \State \hspace{1em} $\mathcal{T}_{\text{reject}}$: Low-quality topics filtered out by the Judge
        \State \hspace{1em} $\mathcal{T}_{\text{final}}$: Final set of high-quality topics used for document retrieval
        % --- Algorithm steps ---
        \State Initialize $\mathcal{T}_{\text{total}}$, $\mathcal{T}_{\text{reject}}, \mathcal{T}_{\text{final}} \gets \emptyset$
        \For{each BISAC subcategory $\mathcal{S}$}
            \State $\mathcal{T}_1 \gets$ Debate$_1$ generates topics
            \State $\mathcal{T}_2 \gets$ Debate$_2$ generates topics
            \State Each Debate model critiques the other's outputs
            \State $\mathcal{T}_{\mathcal{S}} \gets$ Judge filters low-quality topics
            \State $\mathcal{T}_{\text{total}} \gets \mathcal{T}_{\text{total}} \cup \mathcal{T}_1 \cup \mathcal{T}_2$
            \State $\mathcal{T}_{\text{reject}} \gets \mathcal{T}_{\text{reject}} \cup \mathcal{T}_{\mathcal{S}}$
        \EndFor
        \State $\mathcal{T}_{\text{final}} \gets \mathcal{T}_{\text{total}} \setminus \mathcal{T}_{\text{reject}}$
        \ForAll{$t \in \mathcal{T}_{\text{final}}$}
            \State $\mathcal{D}_t \gets$ Top 256 retrieved docs via BM25
            \State $\mathcal{S}_{\text{final}} \gets \mathcal{S}_{\text{final}} \cup$ Aggregate($\mathcal{D}_t$)
        \EndFor
        \end{algorithmic}
    \end{algorithm}

\subsection{Preliminary: BISAC Categories}
    The BISAC classification system is a comprehensive, hierarchical structure consisting of 51 primary categories and approximately 4,500 subcategories, covering nearly all domains of human knowledge. Widely used in the book industry, this system categorizes books by subject, facilitating easier discovery and organization. The BISAC system offers several advantages: (1) \textbf{Comprehensive Coverage}: It spans a wide range of subjects, ensuring representation of nearly all areas of human knowledge, making it a versatile tool for organizing diverse content. (2) \textbf{Hierarchical Structure}: The system is organized hierarchically, allowing for both broad and specific categorization, which aids in efficiently navigating from general topics to more specific subtopics. (3) \textbf{Expert-Developed and Regularly Updated}: Developed by industry experts, the categories are regularly updated to reflect new trends and knowledge areas, ensuring their relevance and accuracy. These advantages make BISAC our chosen basis for organizing long-context data synthesis.

\subsection{Multi-Agent Debate Topic Generation}

    To ensure that the generated topics are both diverse and high-quality, we employ a novel multi-agent debate mechanism that combines generative diversity with discriminative filtering. This approach leverages the complementary strengths of multiple language models: two specialized Debate LLMs independently generate candidate topics, while a Judge LLM identifies and filters out those that are redundant or low-quality. The Judge LLM is typically chosen to be slightly weaker than the Debate LLMs—not only because detecting flaws is generally easier than generating new content \cite{kenton2024on}, but also to reduce inference costs during the judging phase. This design effectively balances adequacy and efficiency.

    For each BISAC subcategory, the two Debate LLMs independently generate candidate topics—each accompanied by a brief explanation—based on their interpretation of the subcategory. They then critique each other's outputs using multiple criteria, including relevance, semantic diversity, complementarity, and overall topical quality, and provide persuasive justifications for their evaluations. The Judge LLM reviews all topics and critiques, flags low-quality or overlapping entries, and constructs the final topic set by removing the rejected topics from the combined outputs of the two Debate models. This process fosters greater topical diversity and conceptual richness than single-model generation, while maintaining content quality through lightweight evaluation.    

\subsection{Document Retrieval and Aggregation}
    Once the final collection of topics $T_{\text{final}}$ is generated, we use each topic to retrieve and aggregate relevant documents into long-context training samples. For each topic in our collection, we use a lightweight BM25 retrieval method, via Manticore Search\footnote{\url{https://manticoresearch.com/}}, to retrieve the top 256 relevant documents from the pretraining corpus. These documents are then aggregated with some strategies to create a sample reaching the target length. The aggregation strategies can be simply random shuffling, or recently proposed long-dependency enhancement methods \cite{tian2024untie,gao2025nextlong}. The above process is formulated in Algorithm~\ref{alg:multi-agent}.

 \subsection{Combining with NExtLong}
    To further enhance the long-context capabilities of LiteLong, we integrate it with the recently proposed NExtLong method \cite{gao2025nextlong}. Initially, we randomly select a document from the retrieved set and apply NExtLong's chunking strategy to this document. The document is divided into multiple meta-chunks, each undergoing hard negative mining to identify challenging negative samples. These negatives are then concatenated with the meta-chunks to form an extended document. In the subsequent step, the model is trained on this synthesized long document, focusing on modeling long-range dependencies by distinguishing the meta-chunks across a wide range of hard negatives. This integration allows LiteLong to leverage NExtLong's strengths in modeling long-range dependencies, further improving the model's ability to handle extensive documents. 
    
    Moreover, the combination of LiteLong and NExtLong significantly enhances resource efficiency. While NExtLong typically operates directly on the pretraining corpus, such as FineWeb-Edu \cite{lozhkov2024finewebedu} and Cosmopedia V2 \cite{benallal2024smollmcorpus}, requiring substantial resources to build large vector retrieval databases, LiteLong + NExtLong only necessitates building a vector retrieval database for the retrieved samples. This is less than one-fifth the size of the original corpus, greatly reducing resource demands.

\section{Experiments}

\begin{table*}[!tbp]
    \centering
    \small
    \begin{tabular}{lccccccc|cc}
    \toprule
    & \multicolumn{7}{c|}{\textbf{Long-Context Metrics}} & \multicolumn{2}{c}{\textbf{GPU Hours}} \\
    \cmidrule(lr){2-8} \cmidrule(lr){9-10}
    \textbf{Method} & \textbf{Recall} & \textbf{RAG} & \textbf{ICL} & \textbf{Rerank} & \textbf{LongQA} & \textbf{RULER} & \textbf{AVG} & \textbf{Embed (H)} & \textbf{Gen (H)} \\
    \midrule
    \multicolumn{10}{c}{\textbf{Comparison with Baseline Methods}} \\
    \midrule
    Standard & 62.33 & 58.67 & 71.24 & 19.18 & 28.99 & 76.68 & 52.85 & 0 & 0 \\
    KNN & 64.24 & 56.00 & 60.28 & 18.77 & 32.27 & 74.30 & 50.97 & 617 & 0 \\
    ICLM & 64.04 & 54.48 & 72.36 & 14.04 & 28.17 & 69.14 & 50.37 & 617 & 0 \\
    Quest & 69.13 & 57.47 & 72.08 & 22.35 & \textbf{33.82} & 76.63 & 55.25 & 0 & 806 \\
    LiteLong & \textbf{83.23} & \textbf{60.43} & \textbf{80.12} & \textbf{30.73} & 33.01 & \textbf{83.88} & \textbf{61.90} & 0 & 6 \\
    \midrule
    \multicolumn{10}{c}{\textbf{Combined with Long-dependency Enhancement Method}} \\
    \midrule
    NExtLong & 82.56 & \textbf{60.91} & \textbf{81.76} & 31.47 & \textbf{37.30} & 81.50 & 62.58 & 928 & 0 \\
    LiteLong+NExtLong & \textbf{82.93} & 60.81 & 80.12 & \textbf{33.68} & 36.97 & \textbf{83.73} & \textbf{63.04} & 170 & 6 \\
    \bottomrule
    \end{tabular}
    \caption{Results on the HELMET and RULER benchmarks, along with GPU resource consumption. The best long-context evaluation results are shown in \textbf{bold}. GPU hours are reported separately for document embedding and topic generation.}
    \label{tab:main_results}
\end{table*}

In this section, we first describe our experimental setup, then present our main results and comparisons with existing methods. Finally, we compare the resource consumption of LiteLong with other methods.

\subsection{Experimental Setup}
    \paragraph{Datasets}
    We utilized two primary datasets for our experiments: FineWeb-Edu \cite{lozhkov2024finewebedu} and Cosmopedia V2 \cite{benallal2024smollmcorpus}. \textbf{FineWeb-Edu} is a curated subset of web content that focuses on educational resources, scholarly articles, and instructional content. \textbf{Cosmopedia V2} serves as a comprehensive knowledge corpus that spans multiple domains, with a strong emphasis on factual accuracy and depth.%These datasets were selected to ensure fair comparisons, as they are also used by the baseline methods reproduced in this work.

    \paragraph{Multi-agent LLMs}
    For the Debate LLMs, we use Qwen2.5-7B \cite{qwen2.5} and 
    Mixtral-8x7B-v0.1 \cite{jiang2024mixtral} as the default models, and for the Judge LLM, we use a smaller model, Gemma3-1B \cite{team2025gemma}.This configuration reflects our efficiency-oriented design: each model in the system is selected to be as lightweight as possible while still sufficiently capable for its respective role.

    \paragraph{Baseline Methods}
    We compare LiteLong with the following baseline methods: \textit{Standard} \cite{ouyang2022training} (random concatenation of documents to reach 128K context length), \textit{KNN} \cite{guu2020retrieval, levine2022the}(implementation of the K-Nearest Neighbors approach from Quest \cite{gao2025quest}), \textit{ICLM} \cite{shi2024incontext} (implementation of the In-Context Learning approach from ICLM \cite{shi2024incontext}), and \textit{Quest} (implementation of the query-centric approach from Quest \cite{gao2025quest}). We also combine LiteLong with a recently proposed long-dependency enhancement method, \textit{NExtLong}, with the implementation from NExtLong \cite{gao2025nextlong}. Baseline method details are shown in Appendix G. All methods synthesize approximately 4 billion tokens of training data to ensure fair comparison.

    \paragraph{Training Hyperparameters}
    We fine-tune LLaMA-3-8B as our base model using each of the datasets described above. Training is performed with a learning rate of 4e-5 (using cosine decay), a batch size of 32 samples (128K tokens each), and 1,000 training steps. We apply a weight decay of 0.1 and use bfloat16 precision. For combined methods such as LiteLong+NExtLong, we first apply the LiteLong process to generate topically coherent document sets, followed by the respective transformation (e.g., chunking in NExtLong).

    \paragraph{Evaluation Benchmarks} 
    We conduct evaluations on the HELMET \cite{yen2025helmet} and RULER \cite{hsieh2024ruler} benchmarks, which are specifically designed to assess long-context understanding across multiple dimensions such as recall, retrieval, reasoning, and comprehension.

\subsection{Main Results}

    Table \ref{tab:main_results} presents the results on the HELMET benchmark \cite{yen2025helmet}, RULER benchmark \cite{hsieh2024ruler} and the resource consumption comparison for different methods.
    \paragraph{Experimental Results Analysis}
    LiteLong achieves the highest average score among all baseline methods, reaching 61.90 across long-context metrics. This represents a significant improvement of 6.65 points over Quest (55.25), the second best baseline. LiteLong demonstrates particularly strong performance on the Recall task (83.23) and the RULER task (83.88), highlighting the effectiveness of our topic-based organization in improving the model’s ability to retain and utilize relevant information from extended contexts.
    
    When combined with long-dependency enhancement methods, LiteLong+NExtLong achieves performance exceeding NExtLong alone (63.04 vs. 62.58 average score), with LiteLong+NExtLong showing particular strengths in Rerank (33.68) and RULER (83.73) metrics. These results suggest that LiteLong's topic-based organization complements techniques focused on attention patterns or negative examples by addressing different aspects of long-context modeling.
    \paragraph{Resource Consumption Comparison}
    LiteLong demonstrates exceptional efficiency advantages over competing methods. While Quest demands considerable computational resources (806 GPU hours for generation), LiteLong delivers superior performance with significantly reduced requirements (only 6 GPU hours for generation). This efficiency advantage is further demonstrated in the combined setting: while NExtLong alone requires 928 GPU hours for embedding, integrating LiteLong reduces the overall cost to just 176 GPU hours (170 hours for embedding and 6 hours for generation). This improvement stems from the fact that NExtLong typically constructs retrieval indices over the entire corpus (e.g., FineWeb-Edu and Cosmopedia V2), whereas LiteLong+NExtLong only requires indexing the much smaller set of documents already retrieved and organized by LiteLong—less than one-fifth the size of the original corpus—thereby substantially reducing embedding and indexing costs. This significant reduction in total resource consumption highlights LiteLong as a practical and efficient solution for long-context modeling, enabling state-of-the-art(SOTA) performance without incurring the significant computational overhead of existing methods.

\section{Ablation Studies}

    In this section, we conduct in-depth ablation studies to analyze the impact of different components and design choices in our approach. Unless otherwise specified, all experiments are conducted with a 128k context length for training, using a mixed dataset composed of FineWeb-Edu \cite{lozhkov2024finewebedu} and Cosmopedia V2 \cite{benallal2024smollmcorpus}.

    \subsection{Effectiveness of BISAC Categories}

    \begin{table}[!tbp]
        \centering
        \small
        \begin{tabular}{lc}
        \toprule
        \textbf{Approach} & \textbf{Average} \\
        \midrule
        GPT4-o Categories & 59.94 \\
        BISAC Categories & \textbf{61.90} \\
        \bottomrule
        \end{tabular}
        \caption{BISAC categories achieve better performance than automatically generated categories, evaluated on the HELMET and RULER benchmarks.}
        \label{tab:bisac_effect}
    \end{table}

    To explore the effectiveness of BISAC categories, we utilize the leading model GPT-4o to synthesize a category system and compare it with the BISAC standard. As shown in Table~\ref{tab:bisac_effect}, data synthesized using the BISAC system achieves the highest average scores on the HELMET \cite{yen2025helmet} and RULER \cite{hsieh2024ruler} benchmarks, reaching an average score of 61.90—surpassing the category system automatically generated by large models. These results suggest that leveraging authoritative, structured external knowledge systems (such as BISAC) can lead to more comprehensive and diverse topic coverage in long-context data synthesis, thereby enhancing the long-text understanding capabilities of downstream models. We believe that as LLMs continue to evolve, automatic classification systems will become increasingly accurate and comprehensive, potentially surpassing BISAC in the future.

\subsection{Effectiveness of Multi-agent Debate Mechanism}

    \begin{table}[!tbp]
        \centering
        \small
        \begin{tabular}{lc}
        \toprule
        \textbf{Method} & \textbf{Average} \\
        \midrule
        LiteLong & \textbf{61.90} \\
        w/o Multi-agent Debate Mechanism & 61.45 \\
        \bottomrule
        \end{tabular}
        \caption{Performance improvements from the multi-agent debate mechanism, evaluated on the HELMET and RULER benchmarks.}
        \label{tab:multi_debate_effect}
    \end{table}

    We further investigate the impact of our multi-agent debate mechanism by comparing it with a \textbf{Direct} approach that employs a single LLM to generate topics without any refinement process. As shown in Table~\ref{tab:multi_debate_effect}, our multi-agent debate mechanism yields substantial improvements over direct generation. It achieves a 0.45-point improvement on average across the HELMET \cite{yen2025helmet} and RULER \cite{hsieh2024ruler} benchmarks, highlighting the effectiveness of the multi-agent debate mechanism. This suggests that the collaborative debate process produces topics of significantly higher quality and diversity than those generated by a single model.

\subsection{Impact of Topic Scaling}
   
    \begin{figure}[!tbp]
      \centering
      \includegraphics[width=0.9\columnwidth]{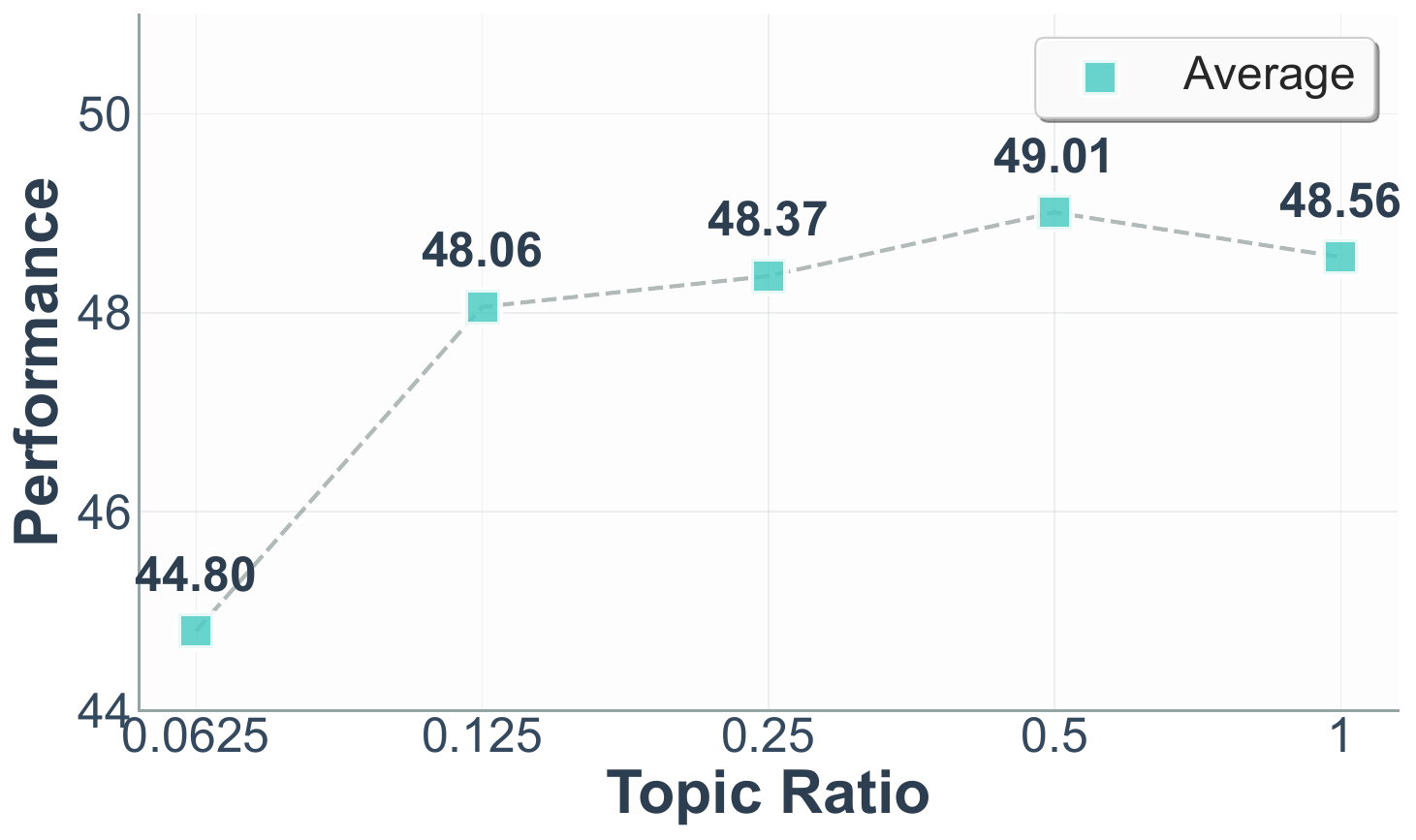}
      \caption{Performance across topic count configurations on the HELMET and RULER benchmarks. The 0.5× configuration (18,720 topics) achieves the highest average score (49.01), with performance degrading as topic count decreases further.}
      \label{fig:scaling_impact}
    \end{figure}

    To understand how the number of topics affects performance, we vary the number of topics in LiteLong while keeping the total token count constant at 2B. As shown in Figure~\ref{fig:scaling_impact}, model performance first improves with an increasing number of topics, but begins to decline once the topic count becomes too high. This trend suggests that it is not merely the number of topics, but the diversity and quality of the topic set that play a crucial role in long-context understanding. Excessive topic granularity may lead to redundancy or reduced per-topic depth, ultimately hindering performance.

\subsection{Impact of Data Source Selection}

    \begin{table}[!tbp]
        \centering
        \small
        \begin{tabular}{lc}
        \toprule
        \textbf{Data Source} & \textbf{Average} \\
        \midrule
        Mix Data & \textbf{61.90} \\
        Cosmopedia V2 Only & 52.32 \\
        FineWeb-Edu Only & 61.29 \\
        \bottomrule
        \end{tabular}
        \caption{Impact of data sources on HELMET performance and RULER performance.}
        \label{tab:data_source_effect}
    \end{table}

    We evaluate the impact of using different data sources for document retrieval. As shown in Table~\ref{tab:data_source_effect}, the mixed dataset (FineWeb-Edu + Cosmopedia V2) achieves the highest average performance (61.90) across the HELMET and RULER benchmarks, outperforming both individual sources. FineWeb-Edu alone yields strong results (61.29), indicating that its diverse writing styles and rich educational content are well-suited for long-context understanding. In contrast, Cosmopedia V2 alone results in a substantially lower score (52.32), suggesting that while curated content may offer high quality, it lacks the variability needed to generalize across complex, long-context tasks. These findings highlight that combining diverse data sources enhances topic coverage and improves downstream performance by providing both breadth and depth in training data.

\subsection{Impact of Topic Retention Strategies in Multi-agent Debate}

    \begin{figure*}[!tbp]
        \centering
        \begin{subfigure}[b]{0.48\linewidth} 
            \includegraphics[width=0.9\linewidth]{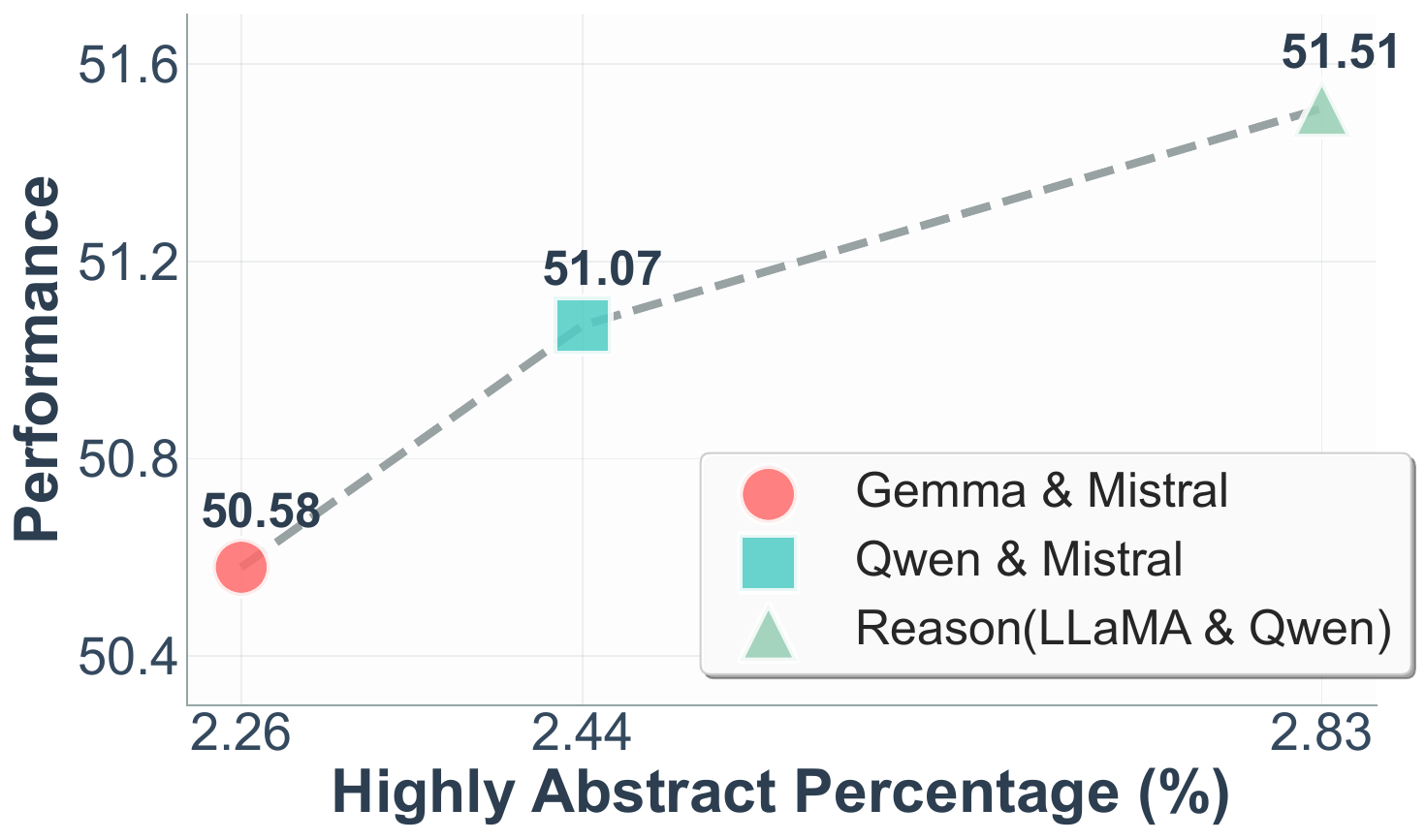}
            \caption{}
            \label{fig:understanding}
        \end{subfigure}
        \begin{subfigure}[b]{0.48\linewidth}
            \includegraphics[width=0.9\linewidth]{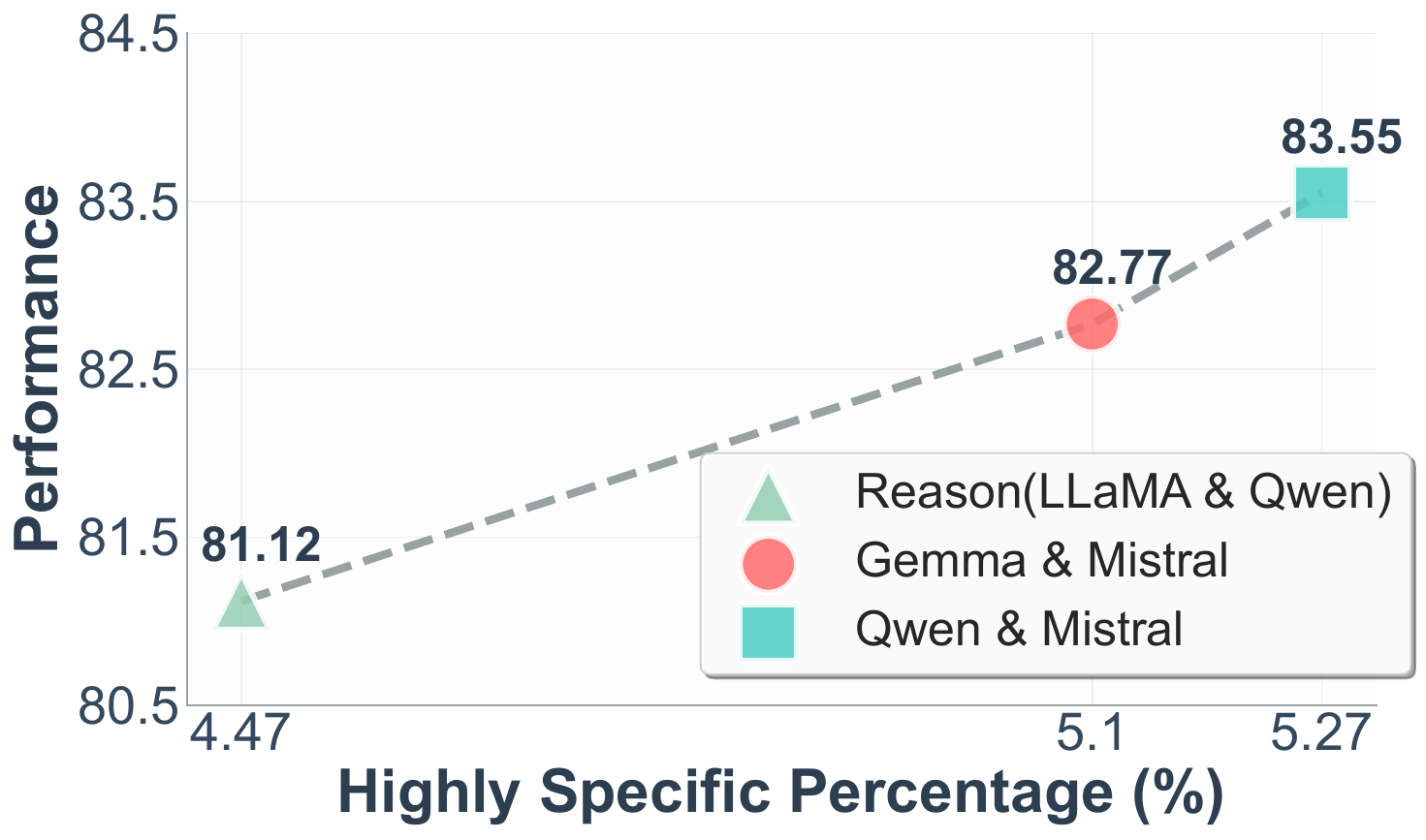}
            \caption{}
            \label{fig:memory}
        \end{subfigure} 
        \caption{
            Impact of Topic Abstraction on Different Task Types. 
            (a) The proportion of highly abstract topics positively correlates with performance in reasoning-oriented tasks (RAG, ICL, Re-rank, LongQA). 
            (b) The proportion of highly specific topics contributes more significantly to performance in memory-intensive tasks (Recall, RULER). 
            \textbf{Model combinations:} 
            \textit{Gemma \& Mistral} refers to Debate LLMs Gemma3-12B and Mixtral-8x7B-v0.1; 
            \textit{Qwen \& Mistral} uses Qwen2.5-7B and Mixtral-8x7B-v0.1; 
            \textit{Reason (LLaMA \& Qwen)} \cite{deepseekai2025} uses Reason-LLaMA-8B and Reason-Qwen2.5-7B \cite{deepseekai2025}.
        }
        \label{fig:topic_distributions}
    \end{figure*}
    
    \begin{table}[!tbp]
        \centering
        \small
        \begin{tabular}{lc}
        \toprule
        \textbf{Retention Strategy} & \textbf{Average Score} \\
        \midrule
        Filter-Reject & \textbf{61.90} \\
        Keep-Accept & 61.34 \\
        Keep-Fixed-K-Accept & 61.54 \\
        \bottomrule
        \end{tabular}
        \caption{
            Comparison of topic retention strategies, all based on decisions from the Judge model. 
            \textbf{Filter-Reject} removes only the topics explicitly rejected by the Judge; 
            \textbf{Keep-Accept} retains all topics the Judge deems acceptable; 
            \textbf{Keep-Fixed-K-Accept} retains a fixed number of accepted topics per subcategory.
        }
        \label{tab:topic_retention_strategies}
    \end{table}
    
    We investigate how different topic retention strategies, guided by the Judge model, influence the final performance. All strategies begin with the union of candidate topics generated by the Debate models:

    \begin{itemize}
    \item \textbf{Filter-Reject}: Remove only those topics explicitly marked as low-quality by the Judge. (Default strategy)
    \item \textbf{Keep-Accept}: Retain all topics the Judge deems acceptable, without imposing a count limit.
    \item \textbf{Keep-Fixed-K-Accept}: Allow the Debate model to directly select a fixed number (e.g., $K = 10$) of high-quality topics.
    \end{itemize}

    As shown in Table~\ref{tab:topic_retention_strategies}, the most effective strategy is \textbf{Filter-Reject}. This approach consistently outperforms strategies where the Judge model either adaptively selects or selects a fixed number of high-quality topics. Specifically, the \textbf{Keep-Fixed-K-Accept} strategy yields slightly lower performance, while the \textbf{Keep-Accept} strategy performs the worst—even falling below the baseline without multi-agent debate. These findings suggest that the Judge model is more effective in identifying low-quality topics than at reliably selecting the optimal ones. Moreover, enforcing a fixed topic count (as in Keep-Fixed-K-Accept) may restrict diversity and coverage. In contrast, the Filter-Reject strategy preserves flexibility in topic selection while eliminating weak content, striking a better balance between quality and coverage.
    
\subsection{Impact of LLM Combinations in Multi-agent Debate}

    We further investigate how different combinations of LLMs used in the multi-agent debate framework influence the characteristics of generated topics. Specifically, we combine various topic generation models (Debate LLMs) with a fixed Judge model (Gemma3-1B), applying a Filter-Reject topic selection strategy. We quantify the abstraction level of topics by computing their average hierarchical depth in WordNet \cite{miller1995wordnet}, classifying topics with a depth less than 3 as highly abstract, and those with a depth greater than 9 as highly specific. Finally, we analyze how these abstraction levels impact downstream performance on different task types.

    We observe that different LLM combinations naturally lead to different distributions of topic abstraction. As shown in Figure~\ref{fig:topic_distributions}(a), topic sets with a greater proportion of abstract topics yield better performance on reasoning-oriented tasks such as Retrieval-Augmented Generation (RAG), In-Context Learning (ICL), Re-ranking, and LongQA. These tasks benefit from broader conceptual coverage and generalization.
    
    In contrast, Figure~\ref{fig:topic_distributions}(b) shows that task performance on memory-intensive evaluations, such as Recall and RULER, is more strongly influenced by specific, narrowly defined topics. Such topics better match the need for detailed content anchoring and information recall.
    
    These findings suggest that the abstraction profile of topic sets—driven by the LLMs used during generation—plays a critical role in shaping downstream task performance. Selecting or mixing LLMs with appropriate abstraction tendencies can thus serve as a tool to tailor training data for different long-context task categories.

\subsection{LiteLong Maintains Performance on Short Tasks}

    \begin{table}[tbp]
        \centering
        \small
        \begin{tabular}{lc}
            \toprule
            
            \textbf{Model} & \textbf{Short Tasks Avg} \\
            \hline
            Base model & 62.05 \\
            LiteLong & 61.92 \\
            \hline
        \end{tabular}
        \caption{Comparison of model performance on short-context tasks, showing that LiteLong maintains comparable performance to the base model. Detailed results for each individual dataset can be found in Appendix B.}
        \label{tab:short_context_results}
    \end{table}

    We also examine whether the LiteLong approach compromises performance on short-context tasks. As shown in Table~\ref{tab:short_context_results}, LiteLong achieves a score of 61.92 on short-context benchmarks, closely matching the performance of the base model of 62.05. This marginal difference demonstrates that LiteLong introduces minimal interference with short-context capabilities. In other words, the integration of long-context data---carefully synthesized through topic-driven document aggregation---does not hinder the model's proficiency in conventional short-sequence NLP tasks. These results confirm that LiteLong effectively preserves short-context generalization while simultaneously strengthening long-context understanding.

\subsection{LiteLong Performs Strongly After Supervised Finetuning}  

    \begin{table}[tbp]
        \centering
        \small
        \begin{tabular}{lcc}
        \toprule
        \textbf{Model} & \textbf{Overall} & \textbf{GPU hours} \\
        \midrule
        SOTA & 30.4 & 928\\
        + LiteLong & \textbf{30.6} & \textbf{176} \\
        \bottomrule
        \end{tabular}
        \caption{ LiteLong performs strongly after supervised finetuning while maintaining its resource-efficient merit, as evaluated on LongBench v2 benchmarks. Result details are presented in Appendix B.}
        \label{tab:sft_results}
    \end{table}

    Finally, we investigate the potential of LiteLong after supervised fine-tuning and evaluate on the LongBench v2 benchmarks \cite{bai2024longbench}. For fair comparison, we follow the supervised fine-tuning procedure from the most recent SOTA method NextLong \cite{gao2025nextlong}. As shown in Table~\ref{tab:sft_results}, combined with LiteLong, the model achieves a competitive score of 30.6, on par with NExtLong’s 30.4, while requiring only 19\% of the GPU resources used by NExtLong. The results indicate that LiteLong's resource efficiency advantage does not come at the expense of downstream performance, demonstrating the enormous potential of LiteLong's applications.

\section{Conclusion}
    This paper introduces LiteLong, a resource-efficient method for synthesizing high-quality long-context training data. Traditional Random Concatenation methods are computationally efficient but lack coherence, while Similarity-Based and Query-Centric approaches improve coherence at the cost of computational overhead. LiteLong employs a BISAC-guided structure and a multi-agent debate mechanism to generate diverse topics and retrieves documents via lightweight BM25 to build 128K‑token sequences. Experiments on the HELMET and RULER benchmarks show that LiteLong delivers competitive performance while keeping computational costs low. Future work may explore other modalities and incorporate more diverse retrieval strategies.

\bibliography{example_paper}
\bibliographystyle{icml2025}

% You can have as much text here as you want. The main body must be at most $8$ pages long.
% For the final version, one more page can be added.
% If you want, you can use an appendix like this one.  

% The $\mathtt{\backslash onecolumn}$ command above can be kept in place if you prefer a one-column appendix, or can be removed if you prefer a two-column appendix.  Apart from this possible change, the style (font size, spacing, margins, page numbering, etc.) should be kept the same as the main body.
%%%%%%%%%%%%%%%%%%%%%%%%%%%%%%%%%%%%%%%%%%%%%%%%%%%%%%%%%%%%%%%%%%%%%%%%%%%%%%%
%%%%%%%%%%%%%%%%%%%%%%%%%%%%%%%%%%%%%%%%%%%%%%%%%%%%%%%%%%%%%%%%%%%%%%%%%%%%%%%

\end{document}